%% file: main.tex
\documentclass[12pt,twoside]{report}

\usepackage{cvpr}
\usepackage{times}
\usepackage{epsfig}

\usepackage[export]{adjustbox}

\usepackage[pagebackref=true,breaklinks=true,letterpaper=true,colorlinks,bookmarks=false]{hyperref}

\cvprfinalcopy % *** Uncomment this line for the final submission

\def\cvprPaperID{****} % *** Enter the CVPR Paper ID here

\ifcvprfinal\fi

\newcommand{\reporttitle}{Image Generation and Recognition (Emotions)}
\newcommand{\reportauthor}{Hanne Carlsson}
\newcommand{\supervisor}{Dimitrios Kollias}

\usepackage[a4paper,hmargin=2.8cm,vmargin=2.0cm,includeheadfoot]{geometry}
\usepackage{textpos}
\usepackage{tabularx,longtable,multirow,subfigure,caption}%hangcaption
\usepackage{fncylab} %formatting of labels
\usepackage{fancyhdr} % page layout
\usepackage{url} % URLs
\usepackage[english]{babel}
\usepackage{amsmath}
\usepackage{amssymb}
\usepackage{graphicx}
\usepackage{dsfont}
\usepackage{epstopdf} % automatically replace .eps with .pdf in graphics
\usepackage{array}
\usepackage{latexsym}

\usepackage{subfiles}

\usepackage[all]{hypcap}

\usepackage{multicol}
\usepackage{booktabs}
\usepackage{listings}

\def\@makechapterhead#1{%
  \vspace*{10\p@}%
  {\parindent \z@ \raggedright \sffamily
    \interlinepenalty\@M
    \Huge\bfseries \thechapter \space\space #1\par\nobreak
    \vskip 30\p@
  }}

\def\@makeschapterhead#1{%
  \vspace*{10\p@}%
  {\parindent \z@ \raggedright
    \sffamily
    \interlinepenalty\@M
    \Huge \bfseries  #1\par\nobreak
    \vskip 30\p@
  }}

\allowdisplaybreaks

\date{May 2019}

\begin{document}

% load title page
% Last modification: 2015-08-17 (Marc Deisenroth)
\begin{titlepage}

\newcommand{\HRule}{\rule{\linewidth}{0.5mm}} % Defines a new command for the horizontal lines, change thickness here

%----------------------------------------------------------------------------------------
%	LOGO SECTION
%----------------------------------------------------------------------------------------

\includegraphics[width = 4cm]{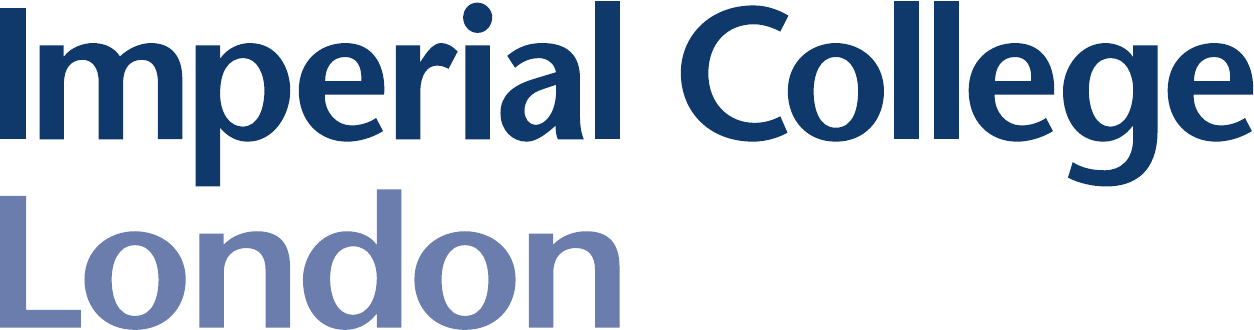}\\[0.5cm] 

\center % Center remainder of the page

%----------------------------------------------------------------------------------------
%	HEADING SECTIONS
%----------------------------------------------------------------------------------------

\textsc{\Large Imperial College London}\\[0.5cm] 
% \textsc{\large Department of Computing}\\[0.5cm] 
\textsc{\large {Independent Study Option}}\\[0.5cm] 

%----------------------------------------------------------------------------------------
%	TITLE SECTION
%----------------------------------------------------------------------------------------
\vspace{2.5cm}
\HRule \\[0.4cm]
{ \huge \bfseries \reporttitle}\\ % Title of your document
\HRule \\[1.5cm]
\vspace{2cm}
%----------------------------------------------------------------------------------------
%	AUTHOR SECTION
%----------------------------------------------------------------------------------------

\begin{minipage}{0.4\textwidth}
\begin{flushleft} \large
\emph{Author:}\\
\reportauthor % Your name
\end{flushleft}
\end{minipage}
~
\begin{minipage}{0.4\textwidth}
\begin{flushright} \large
\emph{Supervisor:} \\
\supervisor % Supervisor's Name
\end{flushright}
\end{minipage}\\[4cm]

%----------------------------------------------------------------------------------------
%	FOOTER & DATE SECTION
%----------------------------------------------------------------------------------------
% \vfill % Fill the rest of the page with whitespace
% Submitted in partial fulfillment of the requirements for the MSc degree in
% \degreetype~of Imperial College London\\[0.5cm]

\makeatletter
\@date 
\makeatother

\end{titlepage}

% page numbering etc.
\pagenumbering{roman}
\clearpage{\pagestyle{empty}\cleardoublepage}
\setcounter{page}{1}
\pagestyle{fancy}

%%%%%%%%%%%%%%%%%%%%%%%%%%%%%%%%%%%%
\begin{abstract}
Generative Adversarial Networks (GANs) were proposed in 2014 by Goodfellow et al.~\cite{original-gan}, and have since been extended into multiple computer vision applications. This report provides a thorough survey of recent GAN research, outlining the various architectures and applications, as well as methods for training GANs and dealing with latent space. This is followed by a discussion of potential areas for future GAN research, including: evaluating GANs, better understanding GANs, and techniques for training GANs. The second part of this report outlines the compilation of a dataset of images `in the wild' representing each of the 7 basic human emotions, and analyses experiments done when training a StarGAN~\cite{stargan} on this dataset combined with the FER2013~\cite{fer2013} dataset.
\end{abstract}

\cleardoublepage
%%%%%%%%%%%%%%%%%%%%%%%%%%%%%%%%%%%%
% \section*{Acknowledgments}
% Comment this out if not needed.

% \clearpage{\pagestyle{empty}\cleardoublepage}

%%%%%%%%%%%%%%%%%%%%%%%%%%%%%%%%%%%%
%--- table of contents
\fancyhead[RE,LO]{\sffamily {Table of Contents}}
\tableofcontents

\clearpage{\pagestyle{empty}\cleardoublepage}
\pagenumbering{arabic}
\setcounter{page}{1}
\fancyhead[LE,RO]{\slshape \rightmark}
\fancyhead[LO,RE]{\slshape \leftmark}

%%%%%%%%%%%%%%%%%%%%%%%%%%%%%%%%%%%%
\chapter{Introduction}\label{intro}
\subfile{Sections/1_intro}

%%%%%%%%%%%%%%%%%%%%%%%%%%%%%%%%%%%%
\chapter{GAN Architectures}\label{gan-architechtures}
\subfile{Sections/2_gan-architectures}

%%%%%%%%%%%%%%%%%%%%%%%%%%%%%%%%%%%%
\chapter{Training GANs} \label{training-gans}
\subfile{Sections/3_training-gans}

%%%%%%%%%%%%%%%%%%%%%%%%%%%%%%%%%%%%
\chapter{Structure of Latent Space} \label{latent-space}
\subfile{Sections/4_latent-space}

%%%%%%%%%%%%%%%%%%%%%%%%%%%%%%%%%%%%
\chapter{Applications of GANs} \label{gan-applications}
\subfile{Sections/5_gan-applications}

%%%%%%%%%%%%%%%%%%%%%%%%%%%%%%%%%%%%
\chapter{Discussion on GANs} \label{gan-discussion}
\subfile{Sections/6_discussion}

%%%%%%%%%%%%%%%%%%%%%%%%%%%%%%%%%%%%
\chapter{Emotion Dataset} \label{emotion-database}
\subfile{Sections/7_emotion-dataset}

%%%%%%%%%%%%%%%%%%%%%%%%%%%%%%%%%%%%
\chapter{Implementation} \label{implementation}
\subfile{Sections/8_implementation}

%%%%%%%%%%%%%%%%%%%%%%%%%%%%%%%%%%%%
\chapter{Conclusion}
\subfile{Sections/9_conclusion}

{\small
\bibliographystyle{ieee_fullname}
\bibliography{egbib}
}

\end{document}

%% file: Sections/1_intro.tex
Generative Adversarial Networks (GANs) are a fairly recent addition to deep learning research, and have advanced generative image modeling dramatically~\cite{big-gan} since they were proposed by Goodfellow et al.~\cite{original-gan} in 2014. GANs are applicable for both semi-supervised and unsupervised learning tasks~\cite{gan-survey}, and they have achieved impressive results in various image generation tasks, such as: image-to-image synthesis, text-to-image synthesis, and image super-resolution~\cite{sa-gan}.
\\ \\
There are many different types of GANs (see Chapter \ref{gan-architechtures}), but they all consist of two simultaneously trained models: a generator and a discriminator. The generator captures the distribution of the data and tries to generate a plausible image from that distribution, while the discriminator estimates the probability of the generated image being from the training data rather than made by the generator~\cite{original-gan}. The two are in competition with each other, with the generator learning to fool the discriminator and the discriminator learning to tell real and fake images apart. Formally, the objective of GANs is to find the Nash equilibrium to the following two player minimax problem with value function $V(D,G)$~\cite{original-gan}: 

\begin{equation} \label{eq:minimax}
    \min_G\max_DV(D,G)=\mathbb{E}_{\Vec{x}\thicksim p_{data}(\Vec{x})}[\log D(\Vec{x})] + \mathbb{E}_{\Vec{z}\thicksim p_{\Vec{z}}(\Vec{z})}[\log (1-D(G(\Vec{z})))]
\end{equation}

The generator defines a probability distribution $p_g$ of samples $G(\Vec{z})$ obtained from a noise distribution $\Vec{z}\thicksim p_\Vec{z}$. Goodfellow et al.~\cite{original-gan} proved that this minimax game has a global optimum for $p_g=p_{data}$, when the discriminator is unable to differentiate between the training data and generator distributions. The generator never sees the real training images, and instead learns the distribution of them through its interaction with the discriminator~\cite{gan-survey}. 

\section{Project Aim} 
This project aimed to explore current GAN research and understand the trends and advancements made since their proposal in 2014. The first part of this project involved conducting an extensive literature review of GAN research to date (see Chapters \ref{gan-architechtures}-\ref{gan-applications}), and then discussing the apparent research trends in the context \cite{kollias12} of open areas of research for future work (see Chapter \ref{gan-discussion}).
\\ \\
The second part of this project involved implementing and applying GANs. First, a dataset of images `in the wild' displaying the 7 basic emotions was compiled (see Chapter \ref{emotion-database}), and then these were combined with the FER2013\cite{fer2013} dataset to train a StarGAN~\cite{stargan} for image-to-image translation of emotions (see Chapter \ref{implementation}).

%% file: Sections/2_gan-architectures.tex
Since GANs were introduced in 2014, many variants have been proposed for different applications. This chapter aims to walk through some of the key differences in various GAN architectures.

\section{Conditional GANs}
\begin{figure}
    \centering
    \includegraphics[scale=0.4]{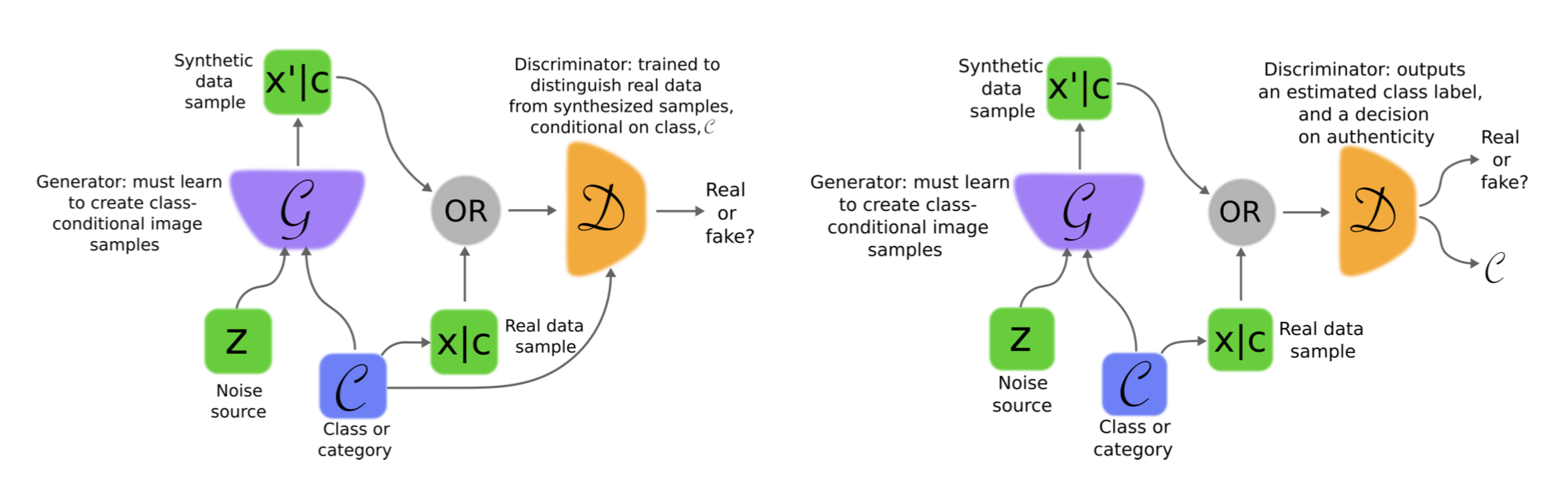}
    \caption{The image on the left is the conditional GAN proposed by Mirza et al.~\cite{conditional-gans}, and the image on the right is the InfoGAN proposed by  Chen et al.~\cite{info-gan}. The image is from the GAN survey paper by Creswell et al.~\cite{gan-survey}.}
    \label{fig:cGAN-infoGAN}
\end{figure}

Conditional GANs were introduced by Mirza et al. as a way to condition the model and be able to direct what it generates~\cite{conditional-gans}. This is done by feeding the data to condition on to both the generator and discriminator as an additional input layer~\cite{gan-survey} (see Figure \ref{fig:cGAN-infoGAN}). An advantage of using conditional GANs is that it enables better representations for one-to-many mappings, meaning conditioning on a single class (eg. Dog) can generate multiple types of dogs with various colours and features. 
\\ \\
The Information Maximising GAN (InfoGAN) proposed by Chen et al.~\cite{info-gan} in 2016 is a completely unsupervised adaptation of the Conditional GAN where the discriminator estimates both if an image is real or fake, and what its class label is~\cite{gan-survey} (see Figure \ref{fig:cGAN-infoGAN}). By adding an information regularisation term to the GAN minimax equation (Equation \ref{eq:minimax}), the InfoGAN enforces high mutual information\footnote{Mutual information measures the amount of information that is learned from one random variable $Y$ about another random variable $X$: $\mathbb{I}(X;Y)=H(X)-H(X|Y)=H(Y)-H(Y|X)$~\cite{info-gan}.} between the condition and the generator distribution~\cite{info-gan}. Despite being able to successfully learn interpretable representations on difficult datasets (eg. CelebA~\cite{celeba}), the InfoGAN barely adds any computational cost to the original GAN.
\\ \\
In 2018 a projection based approach to condition information in the discriminator was proposed by Miyato et al.~\cite{cgan-with-projection}. Instead of concatenating the conditioning label with the training data~\cite{conditional-gans}, their paper instead proposes taking the inner product between the embedded condition vector and the feature vector~\cite{cgan-with-projection}. An advantage of using this projection method in the discriminator is that it does not suffer from mode collapse\footnote{Mode collapse is when the discriminator stops causing the generator outputs to be dissimilar, resulting in the generator only outputting very similar looking images, with no diversity~\cite{training-gans}.}, like previous conditional methods do, and it helps increase the diversity within each class. 
\\ \\
Multiple GAN variants make use of conditional GANs as part of their overall architecture, some of these include: Text to Image Synthesis~\cite{text-to-image}, GAWWN~\cite{gawwn}, pix2pix~\cite{pix2pix}, StackGAN~\cite{stack-gan}, and StarGAN~\cite{stargan}.

\section{Convolutional GANs}\label{convGANs}
Convolutional neural networks (CNNs) are very well suited to solving image related problems, requiring fewer parameters than a normal feed-forward neural network \cite{kollia2009interweaving} and therefore being much faster to train~\cite{cnn}. Convolutional GANs are able to make use of the advantages of CNNs in GANs. 
\\ \\
In 2015 Radford et al.~\cite{dcgan} proposed the Deep Convolutional GAN (DCGAN), where both the discriminator and generator are deep CNNs. Their approach uses the all convolutional net~\cite{all-cnn} in both the generator and discriminator which replaces pooling with strided convolutions, allowing down-sampling and up-sampling to be learned during training~\cite{gan-survey}. They also apply batch normalisation~\cite{batch-norm} everywhere (except for the generator output layer and discriminator input layer) which stabilises training.
\\ \\
The 3D-GAN proposed by Wu et al.~\cite{3d-gan} in 2016 also makes use of an all convolutional net~\cite{all-cnn} in both its generator and discriminator, synthesising 3D images of objects using volumetric convolutions~\cite{gan-survey}. The generator consists of 5 volumetric fully convolutional layers, and given a 200 dimensional noise vector $\Vec{z}$ it outputs a $64\times64\times64$ dimensional image of an object. Their implementation did not make use of conditional GANs and instead they trained one 3D-GAN for each class of objects they were interested in (eg. chairs, or tables). 
\\ \\
A few months later in 2016 an approach for image-to-image translation, called pix2pix, was proposed by Isola et al.~\cite{pix2pix}. The pix2pix model is able to generate photo-realistic images from label maps, reconstruct objects from edge maps, and add colour to black and white images~\cite{pix2pix}. This model uses a U-net\cite{u-net} architecture for the generator, in order to avoid the bottleneck caused by using an encoder-decoder network, and a convolutional PatchGAN\cite{patch-gan} classifier for the discriminator, discriminating on a patch-level rather than on an image-level~\cite{pix2pix}. They use a $70\times70$ PatchGAN to force sharp photo-realistic outputs, after finding that increasing the patch size beyond this lowers the overall output quality. Their framework is not application specific, but instead makes use of a conditional GAN architecture to be able to deal with numerous image-to-image translation problems. 
\\ \\
Numerous other GAN variants make use of convolutional networks as part of their architecture, some of these include: InfoGAN~\cite{info-gan}, GAWWN~\cite{gawwn}, StackGAN~\cite{stack-gan}, CycleGAN~\cite{cycle-gan}, Progressive Growing of GANs~\cite{growing-of-gans}, StarGAN~\cite{stargan}, BicycleGAN~\cite{bicycle-gan}, SAGAN~\cite{sa-gan}, BigGAN~\cite{big-gan}, and StyleGAN~\cite{style-gan}.

\begin{figure}
    \centering
    \includegraphics[scale=0.4]{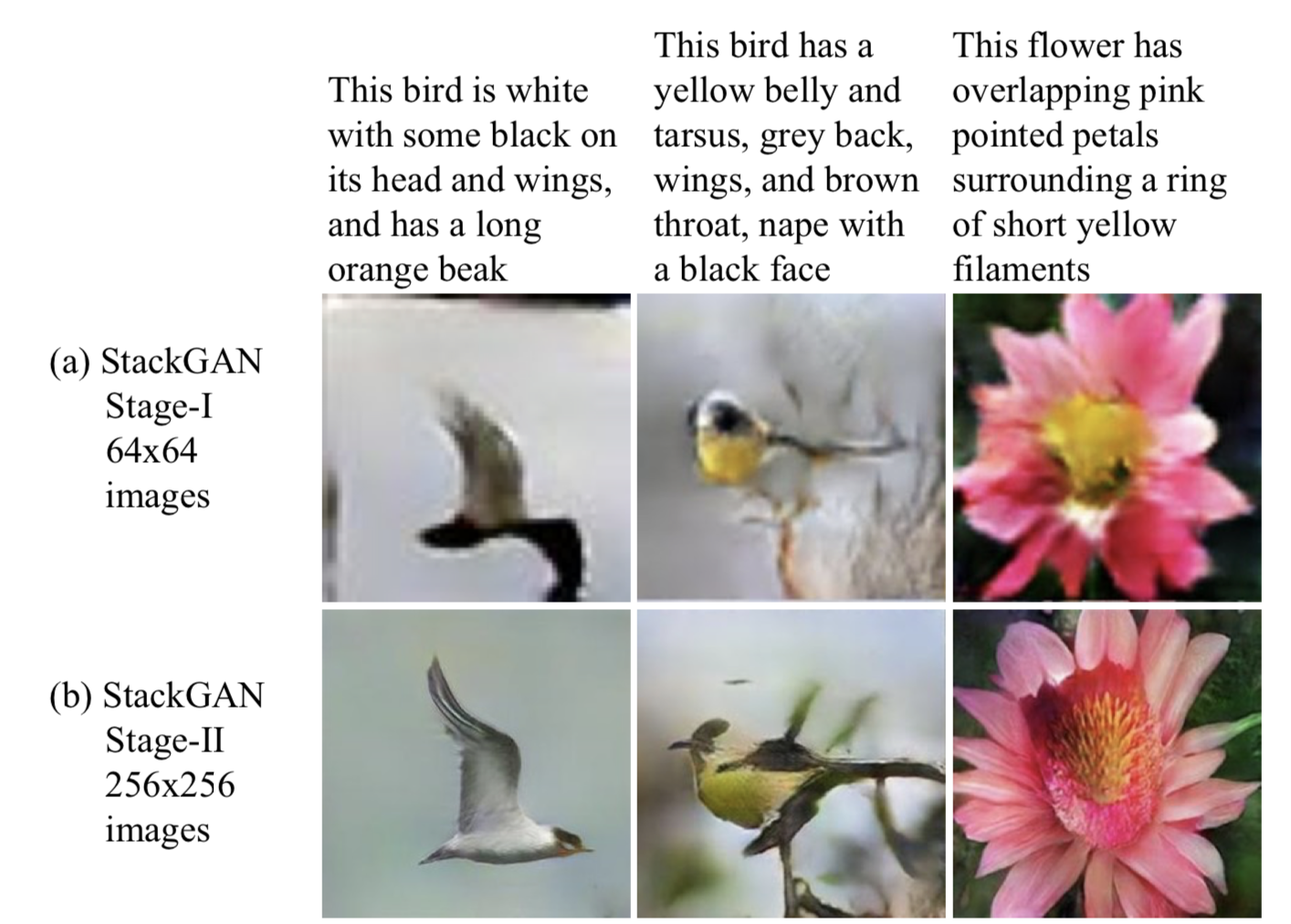}
    \caption{The $64\times64$ images produced by the Stage-I GAN, and the final $256\times256$ images produced by the Stage-II GAN (and therefore full StackGAN) given the various text descriptions on the top. This image is from the StackGAN paper~\cite{stack-gan}.}
    \label{fig:stackgan}
\end{figure}

\section{StackGAN}\label{stackgan}
The StackGAN, proposed by Zhang et al. in 2016~\cite{stack-gan}, is a two-stage GAN that generates photo-realistic images from text descriptions (see Figure \ref{fig:stackgan}). By decomposing this difficult problem into two easier `stages'\footnote{During training they iteratively train Stage-I GAN for 600 epochs while fixing the Stage-II GAN, and then they fix the Stage-I GAN to train the Stage-II GAN for another 600 epochs~\cite{stack-gan}}, they were able to significantly improve on previous state-of-the-art methods. The first stage (Stage-I GAN) sketches the general outline of what is described in the text, and adds the main background and object colours. The second stage (Stage-II GAN) then takes as input both the low-resolution Stage-I output and the text description, focusing on generating the finer details in the image, making it more photo-realistic and true to the text description. 

\section{Cyclical GANs}\label{cyclical-gans}
\begin{figure}
    \centering
    \includegraphics[scale=0.37]{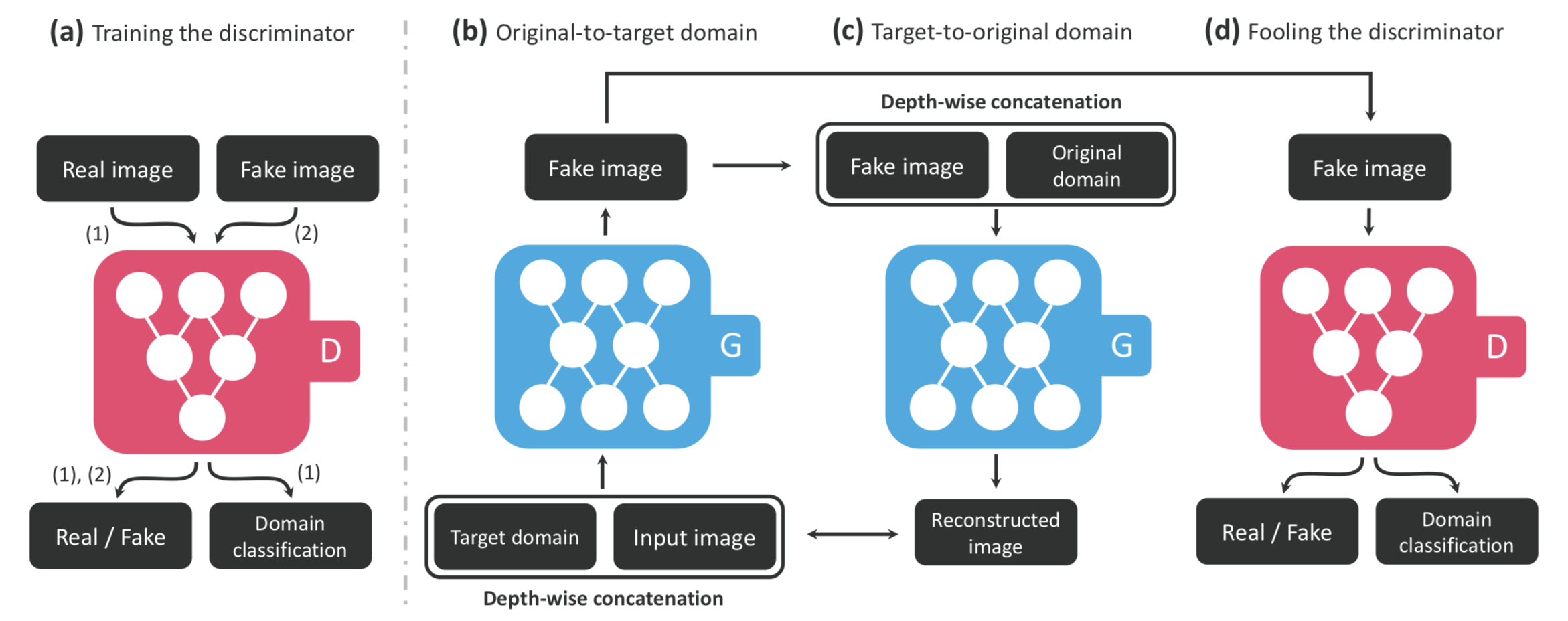}
    \caption{(a) Shows how the StarGAN discriminator is trained, and (b)-(d) show how the StarGAN generator is trained. The image is from the StarGAN paper~\cite{stargan}.}
    \label{fig:starGAN}
\end{figure}

Cyclical GANs make use of a cycle consistency loss which enables translating from one domain to another, and then back again to the starting domain without loss of information. Although cycle consistency loss has been used before, the CycleGAN proposed by Zhu et al.~\cite{cycle-gan} was the first time this was applied to GANs. The CycleGAN enables translating between two different image domains by training on two unordered image collections (one for each domain). This is possible because the CycleGAN assumes an underlying relationship between the two domains, making it possible to train without any paired training data.
\\ \\
A few months after the CycleGAN, the BicycleGAN was proposed by Zhu et al.~\cite{bicycle-gan}. The BicycleGAN model maps input to an output distribution, rather than a deterministic output (like conditional GANs do in previous image-to-image translation models). The BicycleGAN is a combination of a Conditional Variational Autoencoder GAN (cVAE-GAN), which encodes the original input image into latent space, and a Conditional Latent Regressor GAN (cLR-GAN), which gives the generator a randomly drawn latent vector which looks realistic (but not necessarily like the input image) and then uses an encoder to try and recover the latent vector~\cite{bicycle-gan}. This combination of conditional GANs enables the BicycleGAN to produce outputs that are both photo-realistic and diverse in nature.
\\ \\
The StarGAN, proposed by Choi et al.~\cite{stargan}, is unique in that it enables training image-to-image translations between multiple domains with only a single generator and discriminator --- unlike previous image-to-image translation models which can only learn relationships between two domains at a time~\cite{stargan}. It only needs one generator because it uses it twice: first to translate the input image to the target domain, then using a cycle consistency loss it reconstructs the input image from the translated image~\cite{stargan} (see Figure \ref{fig:starGAN}). StarGAN can train on multiple datasets with different label sets, because it uses a mask vector\footnote{A mask vector is a $n$-dimensional one-hot vector, where $n$ is the number of datasets.} which enables the model to ignore unspecified labels.

\section{Self-Attention GAN}
The Self-Attention GAN (SAGAN) proposed by Zhang et al.~\cite{sa-gan} in 2018 uses self-attention and long-range dependency modelling to generate images where the objects and scenarios within it are related in a way consistent with realistic images. Self-attention~\cite{self-attention1,self-attention2} (or intra-attention) finds the relationship between different parts of a sequence in order to represent and understand each part of the sequence. The SAGAN is the first time self-attention has been applied to GANs. Most GAN models use convolutional layers as part of their architecture (see Section \ref{convGANs}) since they are very good at modelling local dependencies, however CNNs cannot efficiently model long-range dependencies on their own. By applying self-attention to both the generator and the discriminator the SAGAN is able to effectively model both local and global dependencies in an image, ensuring that highly detailed features in various parts of the generated image are consistent~\cite{sa-gan}.

%% file: Sections/3_training-gans.tex
In the paper where GANs were first introduced~\cite{original-gan}, the generator network used both Rectified Linear Units (ReLU)~\cite[pg.189]{deep-learning-book} and Sigmoid~\cite{sigmoid} as activation functions, and the discriminator used maxout~\cite{maxout} activations. Dropout~\cite{dropout} was also applied when training the discriminator~\cite{original-gan}. The goal of Goodfellow et al. was never to find the ideal training parameters, however, but rather to introduce GANs and show the theory behind the optimal solution.
\\ \\
Despite an optimal solution theoretically existing, finding it in practice can be very difficult. The main reason for this is because for the original GAN the optimum is a saddle point, rather than a minimum like most other machine learning models. Due to these difficulties, some of the problems which often occur when training GANs are: struggling to have both generator and discriminator converge, mode collapse, and the discriminator converging to zero which gives the generator no useful gradient updates~\cite{gan-survey}. This chapter will discuss some of the findings in later papers that focused on overcoming these issues and improving the GAN training process.

\section{Training Tricks}
The first paper to suggest techniques to improve the training process was the DCGAN paper~\cite{dcgan}, which outlined guidelines for training and constructing both the generator and discriminator. The specific architecture of the DCGAN is outlined in Section \ref{convGANs}, but the addition most related to the training of GANs was the application of batch normalisation throughout the model which helped stabilise learning in deeper networks~\cite{gan-survey}. Radford et al. also showed that using Leaky ReLU~\cite{leakyrelu} in the discriminator, and using tanh~\cite[pg.191]{deep-learning-book} for the output layer of the generator improved performance~\cite{dcgan}. Using tanh allowed the model to learn to saturate quicker and cover the full colour space of the training distribution~\cite{dcgan}.
\\ \\
The DCGAN was trained with mini-batch stochastic gradient descent (SGD) with a mini-batch size of 128, and its weights were initialised from a Gaussian with $\mu=0,\  \sigma=0.02$~\cite{dcgan}. They also used the Adam~\cite{adam} optimiser, which became common practice in GAN training after the release of this paper.
\\ \\
A few months later, Salimans et al.~\cite{training-gans} released a paper proposing further guidelines on stabilising GAN training~\cite{gan-survey}. They observed that gradient descent was not always sufficient when searching for the Nash equilibrium, and introduced five techniques to encourage convergence:~\cite{training-gans}:
\begin{enumerate}
    \item \textbf{Feature matching}: making the objective of the generator be to match the expected value of intermediate discriminator layers. This is effective when the GAN is unstable during training, but there is no guarantee of it working in practice.
    \item \textbf{Minibatch discrimination}: deals with mode collapse by having the discriminator look at a mini-batch of examples rather than a single example, enabling it to tell if the generator is producing the same outputs. This method was found to be superior to feature matching.
    \item \textbf{Historical averaging}: keeps a historical average of parameters in order to penalise parameters that drastically differ from the average.
    \item \textbf{One-sided label smoothing}: changes the discriminator target from 1 to $0.9$ to stop it from becoming overly confident and providing weak gradients.
    \item \textbf{Virtual batch normalisation}: uses the advantages of batch normalisation found by Radford et al.~\cite{dcgan}, but normalises each example based on a  reference batch fixed at the start of training. This is very computationally expensive, and was therefore only applied to the generator network.
\end{enumerate}

Isola et al.~\cite{pix2pix} suggested an alternative method to optimising training by elaborating on the original method by Goodfellow et al.~\cite{original-gan} of alternating between updating the generator and discriminator after each iteration, but when optimising the discriminator they divide the objective by 2 which slows down the rate at which the discriminator learns compared to the generator~\cite{pix2pix}. They also, quite unusually, apply dropout to the generator network at test time, and they apply batch normalisation with the statistics of their test batch rather than a training batch.

\section{Variations in GAN Training Techniques}
\begin{figure}[t]
    \centering
    \includegraphics[scale=0.4]{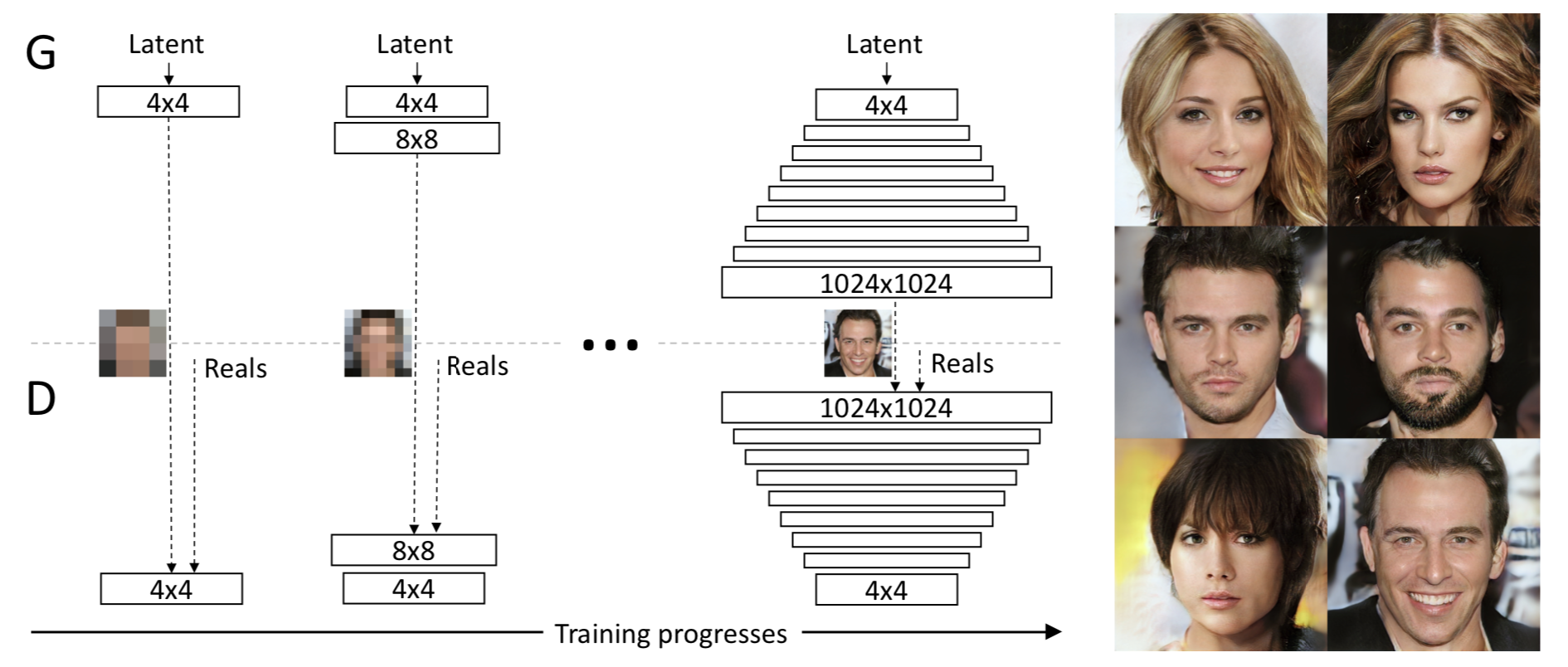}
    \caption{During training the progressively growing GAN begins with the generator and discriminator both having a spatial resolution of $4\times4$. As training advances, they keep adding layers to both networks to increase the resolution. This image is from the paper where this method was introduced~\cite{growing-of-gans}.}
    \label{fig:grow-gan}
\end{figure}

The Wasserstein GAN (WGAN)~\cite{wgan}, proposed by Arjovsky et al., uses a different cost function to previous GANs, namely one derived from an approximation of the Wasserstein distance. The advantages of using this cost function are that it provides more meaningful gradients for updating the generator, the WGAN does not suffer from mode collapse, and it makes the GAN easier to train by avoiding the vanishing gradient problem that other GANs suffer from~\cite{gan-survey}. Another main advantage of the WGAN compared to the original GAN is that the optimum is a minimum rather than a saddle point, like previous GANs. Arjovsky et al. found that using the Adam optimiser did not work for this cost function as it made training very unstable, so instead they used the RMSProp~\cite{rmsprop} optimiser, and decreased the initial learning rate, which helped stabilise training~\cite{wgan}. Training WGANs was improved further by Gulrajani et al.~\cite{wgan-gp} with the proposal of gradient penalty (WGAN-GP) which helped alleviate some of the issues caused by weight clipping in the original WGAN.
\\ \\
Karras et al.~\cite{growing-of-gans} also introduced a new way of training GANs. They progressively increased the resolution of images by adding layers to the generator and discriminator during training (see Figure \ref{fig:grow-gan}), which resulted in more stable training since the majority of training time was spent on smaller low-resolution images that are easier to train on~\cite{growing-of-gans}. They also adapt the suggestion by Salimans et al.~\cite{training-gans} to use minibatch discrimination, resulting in a much simpler approach~\cite{growing-of-gans}:
\begin{enumerate}
    \item Compute standard deviation for every feature (in each spatial location) over the minibatch.
    \item Average estimates over all features and locations.
    \item Replicate and concatenate the value to all spatial locations over the minibatch, resulting in an additional feature map. This layer can be inserted anywhere in the discriminator but they achieved the best results when inserting it towards the end of the discriminator network. 
\end{enumerate}

Another aspect of training that differs for the progressively growing GANs is that instead of using batch normalisation they initialise their weights from $\mathcal{N}(0,1)$ and then scale them at runtime, and they normalise each pixel's feature vector to unit length in the generator after each convolutional layer~\cite{growing-of-gans}. 
\\ \\
A year later (2018) Zhang et al. proposed two techniques to help stabilise training GANs on challenging datasets, which they used for their SAGAN~\cite{sa-gan}. Firstly, they applied spectral normalisation~\cite{spectral-normalisation} to both the generator and discriminator, which allowed for fewer discriminator updates per generator update and reduced the computational training cost~\cite{sa-gan}. Secondly, they used different learning rates for the generator and discriminator, enabling them to get better results in the same wall-clock time\footnote{Wall-clock time is the amount of time it takes to complete a job, which in this case refers to the time it takes to train the GAN.}~\cite{sa-gan}.
\\ \\
Using the SAGAN model as a baseline, the BigGAN proposed by Brock et al.~\cite{big-gan} increased the batch size, width, and depth of the network to train GANs at the largest scale attempted to date~\cite{big-gan}. They found that increasing batch size and width improved performance, presumably because it allowed each batch to cover more modes and thereby provide better gradients, and because increasing the network width increases the number of parameters and thereby the model capacity~\cite{big-gan}. Increasing the depth of the network, however, did not improve performance until they adapted their model to a BigGAN-deep variant which used a slightly different residual block structure than BigGAN~\cite{big-gan}. Scaling up by this amount significantly decreased training stability, which they dealt with in the following ways:

\begin{itemize}
    \item Use a shared embedding (for class embeddings used in conditional batch normalisation layers of the generator) rather than a layer for each embedding. This decreases overall training time and computation.
    \item Add skip-$z$ connections (see Section \ref{inject-latent}).
    \item Truncation trick (see Section \ref{adapt-latent}).
    \item Orthogonal regularisation (see Section \ref{adapt-latent}).
    \item Apply early stopping to deal with training collapse.
\end{itemize}

Brock et al. found that training stability is not dependent on just the generator or the discriminator, but on their interaction throughout training. They also find that increasing training stability by adding regularisation to the discriminator comes at the cost of decreasing the overall training performance, and results in the discriminator overfitting to training data~\cite{big-gan}. The solution which offered the best trade-off was allowing for training collapse later on in training and implementing early stopping when that occurs.

%% file: Sections/4_latent-space.tex
\begin{figure}[]
    \centering
    \includegraphics[scale=0.45]{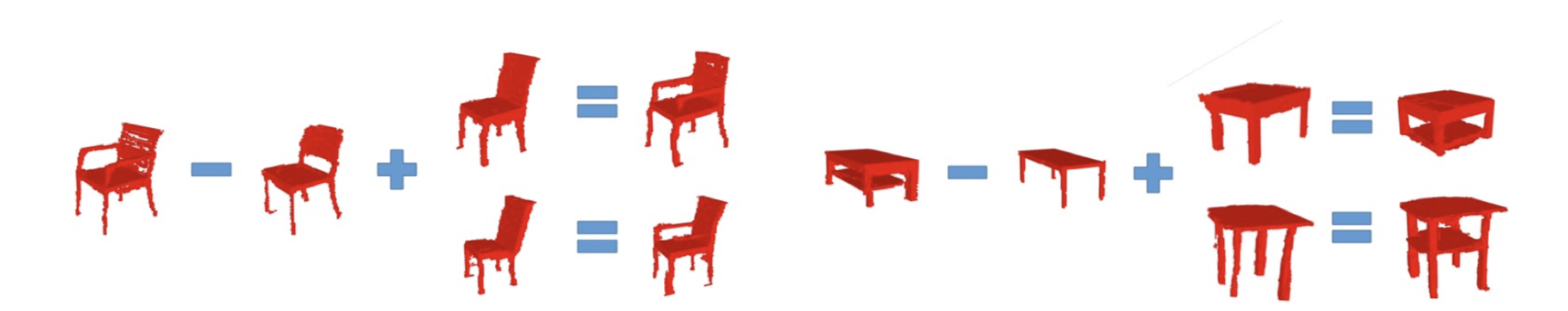}
    \caption{Examples of the kinds of shape arithmetic that can be performed on the latent space of the 3D-GAN. The image is from the 3D-GAN paper~\cite{3d-gan}.}
    \label{fig:shape-arithmetic}
\end{figure}

The generator in all (vanilla) GAN models take a randomly sampled vector from the latent space as input and map it to the domain which it tries to model~\cite{gan-survey}. The latent space has fewer dimensions than the domain space but represents the semantic \cite{kollia2010semantic} structure of the space, in a similar way to word2vec~\cite{word2vec} which represents the semantics of words in $n$\footnote{The dimensionality of word2vec vectors vary depending on how they were trained and what they will be used for, the most common dimensions used are 50, 100, 300, or 500.}-dimensional space. 
\\ \\
A way of exploring this semantic space \cite{kollias10} is to show arithmetic in it. In the word2vec space it is possible to `add' and `subtract' meanings from words in a way that makes sense, eg. king$-$man$+$woman$=$queen. Radford et al.~\cite{dcgan} applied vector arithmetic on face samples, adding/subtracting things like glasses and emotions \cite{kollias6}, but they needed to use the average of three latent vectors in order to obtain stable results. Wu et al.~\cite{3d-gan} showed greater progress with latent space arithmetic by doing shape arithmetic for their 3D-GAN. This allowed them to add or subtract `arm' or `leg' vectors from `table' or `chair' outputs (see Figure \ref{fig:shape-arithmetic}), thus demonstrating the learned representations of the 3D-GAN output~\cite{3d-gan}.
\\ \\
The way the latent space has been modelled and fed to the generator varies between different GAN implementations, depending on the overall goal of the GAN model. In the AlignDRAW GAN proposed by Mansimov et al.~\cite{align-draw} (which generates images from captions) the latent variables are randomly sampled from a normal distribution where the mean and variance are dependent on the previous hidden states of the generative LSTM~\cite{lstm}, rather than drawn from a $\mathcal{N}(0,I)$ distribution~\cite{align-draw}. For the InfoGAN model, Chen et al.~\cite{info-gan} chose to decompose the noise vector into two separate parts: one treated as the noise, and one which targets the semantic features of the distribution \cite{kollias11}. In the pix2pix model, Isola et al.~\cite{pix2pix} do not provide any noise to the generator, and instead apply dropout to several layers throughout the generator network at both training and test time to act as noise in the network.
\\ \\
This chapter will cover some of the methods attempted when dealing with latent space, and the ways latent code is added to generator networks.

\section{Adapting the Latent Space}\label{adapt-latent}
\begin{figure}
    \centering
    \includegraphics[scale=0.48]{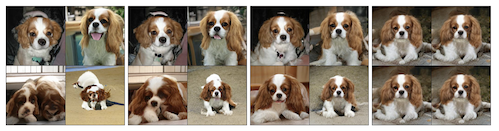}
    \caption{These images show the effects of decreasing the threshold of the Truncation Trick on the BigGAN. From left to right the threshold is $2$, $1$, $0.5$, $0.04$. This image is from the BigGAN paper~\cite{big-gan}.}
    \label{fig:biggan-trunc}
\end{figure}

Zhang et al.~\cite{stack-gan} proposed using \textit{Conditioning Augmentation} for their text-to-image StackGAN. This was an attempt to mitigate previous discontinuity issues in the latent space caused by having limited amounts of data when transforming text embeddings into the latent space~\cite{stack-gan}. Their proposed method randomly samples additional conditioning variables from a Gaussian distribution with a mean and covariance matrix which are both functions of the text embedding. This mechanism helped produce more training pairs than before from the same size dataset.
\\ \\
Zhu et al.~\cite{bicycle-gan} investigated how dimensionality of the latent space can impact the overall model. They found that having too low of a dimensional space may limit the amount of diversity the latent space can represent, whereas having a very high-dimensional space can make sampling difficult since too much information will be encoded in the space. The optimal dimensionality will be dependent on the intended application of the model and the individual dataset used~\cite{bicycle-gan}.
\\ \\
The \textit{Truncation Trick} was proposed by Brock et al.~\cite{big-gan} for the BigGAN, and is also used by Karras et al.~\cite{style-gan} for the StyleGAN (although only for low resolutions to avoid affecting high resolution details). They truncate their $z$ vector by a threshold, and then every value with a magnitude above the chosen threshold is re-sampled. As the threshold value decreases, the Inception Score (IS\footnote{IS is a metric for evaluating GANs that does not penalise a lack of variety but instead rewards precision~\cite{big-gan}. (The higher the score the better)}) increases, and the Fr\'echet Inception Distance (FID\footnote{FID is another commonly used metric for evaluating GANs. It penalises a lack of variety in generated images, but also rewards precision~\cite{big-gan}. (The lower the distance the better)}) initially improves, but then as the threshold approaches zero it worsens to penalise the lack of variety (see Figure \ref{fig:biggan-trunc}). Interestingly they obtain their best overall results when they train their model with a $\mathcal{N}(0,I)$ latent distribution and then sample with the Truncation Trick.
\\ \\
Another `trick' applied by Brock et al.~\cite{big-gan} is the use of \textit{Orthogonal Regularisation}~\cite{orthogonal-reg} to allow the full latent space to map to `good' samples. This aims to minimise the cosine similarity between filters and results in adding smoothness to the models. They find that $60\%$ of their models are responsive to the Truncation Trick when they have been trained with Orthogonal Regularisation~\cite{big-gan}.

\begin{figure}
    \centering
    \includegraphics[scale=0.38]{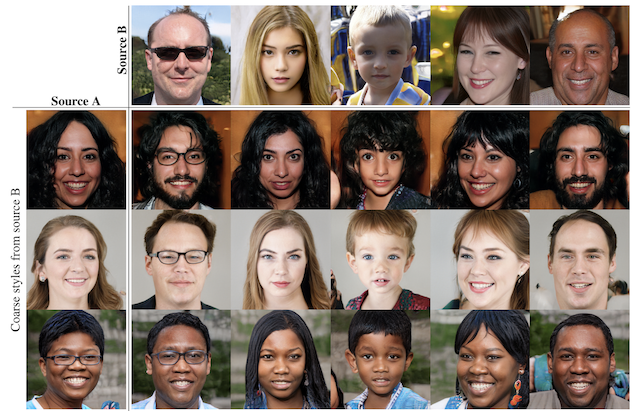}
    \caption{Examples of images generated by applying mixing regularisation and mixing two latent codes at coarse spatial resolutions during training. Aspects such as pose, hair style, and face shape are copied from source B, while the colours and finer facial features are from source A. This image is from the StyleGAN paper~\cite{style-gan}.}
    \label{fig:style-mixingreg}
\end{figure}

Karras et al.~\cite{style-gan} applied \textit{Mixing Regularisation} to their StyleGAN implementation, forcing some images to be generated using two random latent codes instead of just one during training (see Figure \ref{fig:style-mixingreg}). If two latent codes are used, then the generator will randomly switch between the two throughout the network. This stops the network from assuming that adjacent styles are correlated.

\section{Injecting Latent Code}\label{inject-latent}
Zhu et al.~\cite{bicycle-gan} explored two ways of injecting latent code into the generator for their BicycleGAN model: injecting it into the generator's input layer, and injecting it into every intermediate layer of the network~\cite{bicycle-gan}. They obtained similar performance from both methods, showing that for their model it did not make much difference where the latent code was added. 
\\ \\
Brock et al.~\cite{big-gan} also inject latent code into multiple intermediate layers of the generator (skip-\textit{z} connections). They found that for BigGAN and BigGAN-deep this only boosted the overall accuracy by around $4\%$, but that it improved the training speed by a further $18\%$~\cite{big-gan}. They suggest that this is because skip-\textit{z} connections allow the latent space to influence features at different resolutions throughout the generator network~\cite{big-gan}.

%% file: Sections/5_gan-applications.tex
Along with the general advancements of GAN research, discovering new applications for GANs is another active research field~\cite{gan-survey}. The application and goal of a GAN model, as mentioned numerous times throughout this report, has a major impact on its general architecture and training set up. This chapter will highlight some of the broad categories GAN applications fall into and the type of GAN models that fall into each category.

\section{Image Classification}
Image classification is a key part of computer vision research \cite{kollias13,goudelis2013exploring,simou2008image,caridakis2006synthesizing,doulamis1999interactive,tagaris1,tagaris2}, and being able to apply GANs to this is very useful. One way GANs can be used for image classification is as a way to quantitatively assess features extracted from unsupervised learning. Radford et al.~\cite{dcgan} reuse the outputs of the convolutional layers in the discriminator (after training) as a feature extractor and assess their quality by applying a regularised L2-SVM classifier to each feature vector~\cite{gan-survey}. Due to the difficulty in classifying an image as real or fake (generated by a GAN) as GANs improve, using image classification is an important tool in judging the performance of a GAN~\cite{gan-survey}.

\section{Image Synthesis}
\begin{figure}
    \centering
    \includegraphics[scale=0.35]{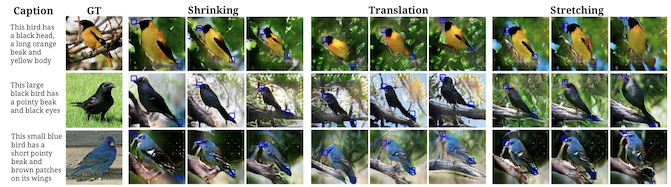}
    \caption{The positioning of the birds are determined by the keypoint coordinates, which enables shrinking, stretching, and translation of the bird in the image. Caption is the text input, GT is the Ground Truth, and the remainder of the images are the generated output based on the keypoint coordinates. This image is from the GAWWN paper~\cite{gawwn}. }
    \label{fig:gawwn}
\end{figure}

One of the most important qualities of a GAN is its ability to generate (often photo-realistic) images \cite{kollias8,kollias9}. This becomes especially useful when the generated image can be conditioned on certain constraints relating to its features~\cite{gan-survey}. Conditioning was first introduced with the Conditional GAN, where the generated image was conditioned on a certain class~\cite{conditional-gans}. This was expanded into `text-to-image' synthesis, where the generated image was conditioned on a short text description.
\\ \\
The AlignDRAW~\cite{align-draw} GAN attempted to solve this problem, but was unable to do so by training an end-to-end model. In 2016 Reed et al.~\cite{text-to-image} proposed the first end-to-end differentiable architecture which conditioned on text descriptions rather than class labels. Their results were not particularly photo-realistic and detailed, but they showed the possibilities of conditioning on natural language, and in the three years since their proposal text-to-image synthesis has made impressive improvements.
\\ \\
Later that same year, Reed et al.~\cite{gawwn} proposed the Generative Adversarial What-Where Network (GAWWN) which was able to condition on where specific parts of the text descriptions should be located in the image, using bounding boxes. They show that they are able to perform shrinking, stretching, and translation of the specific objects by altering the bounding box coordinates, without changing the text or noise variables the model is conditioned on (see Figure \ref{fig:gawwn}).
\\ \\
The StackGAN~\cite{stack-gan} also performed text-to-image synthesis, but split the problem into two `stages', one producing a general low-resolution outline of the text description, and the next improving the resolution of the first image by adding high-resolution details (more details on the specific architecture are given in Section \ref{stackgan}).
\\ \\ 
The quality of images generated by GANs has improved greatly since the GAN was introduced~\cite{original-gan}. Very recently the BigGAN~\cite{big-gan} was proposed, which produced incredibly high resolution and detailed photo-realistic images at a larger scale than previously attempted. This showed the potential possibilities of photo-realistic images being generated from GANs by scaling up.

\section{Image-to-Image Translation}
\begin{figure}
    \centering
    \includegraphics[scale=0.37]{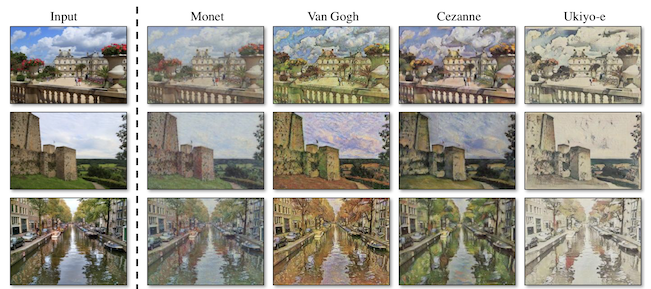}
    \caption{The CycleGAN is able to learn the characteristics of various domains, without training on any specific pairs between them, and use these characteristics to translate an input image into a different domain. Here an input photograph has been translated into the style of various famous painters. This image is taken from the CycleGAN paper~\cite{cycle-gan}. }
    \label{fig:cycle-styletransfer}
\end{figure}

Another common application for GANs is image-to-image translation. This is when an image is given as input and the GAN translates it to a different domain. The first image-to-image translation model with a generic enough approach to be applied to multiple different problems was proposed by Isola et al.~\cite{pix2pix} and nicknamed pix2pix. Their model uses a conditional GAN and they show it works for numerous applications such as adding colour to a black and white image, translating an edge map to a photo-realistic image, translating a image between night and day, and creating a map from aerial photographs. These problems had previously required separate models trained for each specific domain translation~\cite{gan-survey}.
\\ \\
This was later extended by the CycleGAN~\cite{cycle-gan} which introduced a cycle-consistency loss enabling reverse translations without a loss of information. This cyclical approach meant that matching image pairs were no longer needed for training, and instead it was sufficient to train on two individual domains for the GAN to learn the characteristics of each well enough to translate between the two (see Figure \ref{fig:cycle-styletransfer}). Other models which made use of this cycle-consistency loss for image-to-image translation were the BicycleGAN and the StarGAN, and the details of their architecture are discussed in Section \ref{cyclical-gans}.

%% file: Sections/6_discussion.tex
This chapter will discuss the methods commonly used to evaluate GANs, and the advantages and (more importantly) disadvantages of using these, as well as the areas of future work within GAN research where there are still unanswered questions and room for improvement.

\section{Evaluating GANs}\label{evaluating-gans}
As GANs output images, an obvious way of evaluating GANs is to look at the outputs and judge how realistic and reasonable they seem. The issue with this is it is hard to compare various GAN models, or even alterations of the same model, without having a way to quantify how `good' a given model is for a certain task. To date there is no overall metric used to compare all GANs, but there are a few main ones that have been used to compare models with the same applications.
\\ \\
Image-to-Image models are often quantitatively evaluated using the FCN score. FCN stands for Fully-Convolutional Network, and the FCN score uses a classifier to perform semantic segmentation on the output images, and compares this to a ground truth labelled image. The intuition behind this metric is that if the GAN has successfully generated a realistic image, then a classifier should be able to label its semantic components successfully~\cite{cycle-gan}. For Image-to-Image models that also perform photo-to-label translations (eg. CycleGAN~\cite{cycle-gan}) metrics such as per-pixel accuracy, per-class accuracy, and mean class Intersection-Over-Union (Class IOU) are commonly used.
\\ \\
For conditional image synthesis GANs two commonly used metrics are Fr\'echet Inception Distance (FID) and Inception Score (IS). These metrics were briefly mentioned in Section \ref{adapt-latent}, but essentially they are ways to quantitatively  measure the sample quality of a model. They are not perfect measurements, but as they have been used to measure the effectiveness of multiple models, they are useful metrics in that they enable quantitative comparison between models~\cite{big-gan}.
\\ \\
In terms of human evaluation of GAN models, the most common way this is done is by using Amazon Mechanical Turk (AMT). This way the opinions of multiple people can be quantitatively used, which --- although incredibly subjective --- is a valid way of getting a general feel for how good or bad a model is. The most common way AMT is used is that either a single picture is shown at a time and the person has to say whether they believe the image is real or fake (essentially acting as a human discriminator) or multiple pictures are shown and they have to select the real/fake one. These human evaluations are incredibly useful, but they are very hard to use when comparing two models (unless a human evaluation study has been completed to compare the two models).

\section{Areas for Future Work}
GAN research has made great strides since they were first introduced in 2014 by Goodfellow et al.~\cite{original-gan}, but they are still far from perfect and there are multiple areas where future work is needed. The main areas that will be discussed in this section are evaluating GANs, understanding the inner workings of GANs, and training instability.
\\ \\
Regarding the methods for evaluating GANs, as mentioned in Section \ref{evaluating-gans}, although multiple metrics exist all of them are flawed in some way. There is still no one good metric to successfully measure both how well a GAN is performing for its assigned task, and how good the quality of its output is, while also being able to compare multiple models with each other. This is an area of research that is still very open to future work, especially since the current method of using multiple different metrics easily leads to conflicting conclusions regarding the quality of models~\cite{gan-survey}.
\\ \\
Another area for future work is regarding how and why GANs actually work. Part of this is better understanding the latent space and what information it entails. Some of the current methods for manipulating the latent space and using it to control GAN output were discussed in Chapter \ref{latent-space}, but there are still a lot of unknowns regarding how the latent space is actually encoding the information\cite{simou2007fire}. Work was done by Bau et al.~\cite{gan-dissection} to visualise and understand GANs and their inner workings, enabling them to compare the internal representations across various models and datasets as well as be able to manipulate objects within a generated scene. There are, however, still many unknown questions regarding GANs, such as: what is the relationship between layers of a GAN? how can we predict which relationships a GAN will and will not be able to learn?
\\ \\
Finally, as discussed in Chapter \ref{training-gans}, training GANs is still a challenge, and they are very prone to instability and mode collapse. Training GANs requires searching for a Nash equilibrium of the minimax game (Equation \ref{eq:minimax}), but sometimes no equilibrium exists and the GAN is unable to converge to the optimal solution. A lot of work has been done with regards to finding optimal conditions for training, but since they are application dependent, it is still very much an open problem to find ideal training conditions. Brock et al.~\cite{big-gan} trained GANs at the largest scale yet, and found both that their model was very prone to instability during training, and that it achieved incredibly impressive photo-realistic results. This brings about some interesting questions: how far can they be scaled up while still showing a performance improvement? how much scaling up is too much? how much are we able to scale up given our current computing resources?
\\ \\
Being able to better understand and control GANs during training is an area of GAN research that could drastically improve GAN performance as a whole.

%% file: Sections/7_emotion-dataset.tex
As part of this project a database of images representing each of the 7 basic emotions (angry, disgust, fear, happy, neutral, sad, surprise) `in the wild' was compiled. An image is referred to as `in the wild' if it is not posed, and instead taken in a natural environment \cite{kollias1,kollias2,kollias3}. This type of database is quite invaluable in the field of Computer Vision\cite{kollias4,kollias5,kollias15}, because most databases consist of perfectly posed pictures with a natural background, which makes it hard for models trained on those databases to generalise to the real world. By using images `in the wild', or a mixture of posed and `in the wild' images, models are able to better generalise to real world scenarios.
\\ \\
The initial aim was to gather enough images to be able to further categorise each emotion with its intensity, however the size of the final dataset was not large enough for this to seem reasonable. This chapter will detail the process of finding the images used for this database, as well as the pre-processing steps that were done before using the dataset in the Implementation (see Chapter \ref{implementation}).
\section{Finding Images}
The images for this dataset were creative commons licensed images collected from the following websites:
\begin{itemize}
    \item \url{https://pixabay.com/}
    \item \url{https://www.pexels.com/}
    \item \url{https://www.flickr.com/}
    \item \url{https://www.istockphoto.com/}
    \item \url{https://www.freeimages.com/}
    \item \url{https://burst.shopify.com/}
\end{itemize}
The images were downloaded using the Google Chrome extension ``Download All Images'' offered by \url{https://mobilefirst.me}. This enabled multiple images to be downloaded at once, saving immense amounts of time.
\\ \\
Various keywords and phrases were used an in attempt to download as many images as possible for each emotion. This both meant using synonyms for each of the emotions, such as: enraged, annoyed, frustrated, furious, heated, or irritated when looking for images representing the `Angry' class. But also using phrases like ``Angry man'', ``Angry boy'', ``Angry woman'', ``Angry girl'', and ``Angry family''. This meant that more images could be found and that the images collected were quite varied in terms of the ages and number of people present. 
\\ \\
Some emotions were much easier to find images for than others, resulting in quite an in-balance for the final dataset. Compared to other emotion datasets, however, this is the norm since some emotions are simply more represented then others. The easiest emotion to find images for was Happy, and so that is the largest of the categories in the dataset with a total of $559$ images. Then Neutral, Sad, and Angry were all fairly easy to find images for and have a similar amount of representation in the dataset. The remaining emotions (Disgust, Fear, and Surprise), however, were very difficult to find good images for. Few of the websites available for creative commons images have particularly good search engines, and the use of keywords and short phrases had no significant impact when it came to these emotions -- resulting in them all being quite under-represented in the final database.
\\ \\
The total number of images in this newly created database is $1,463$. As this is clearly not enough to use as a training set, it seemed a better idea to combine these images with another emotion dataset in order to have a more reasonable sized training set. The dataset chosen to combine this newly created dataset with was FER2013~\cite{fer2013}, containing a total of $35,887$ images. Despite the overall size of the dataset, it only contains $547$ images for the Disgust class, showing how common the under-representation of that more unusual emotion is. \\ \\
The images in the FER2013 dataset have the same 7 emotion categories as the dataset created throughout this project, and the images are grey-scale crops of faces with dimension $48\times48$. In order to combine the two datasets, some pre-processing was required (Section \ref{pre-process}).

\section{Pre-Processing}\label{pre-process}
\begin{figure}
    \centering
    \includegraphics[scale=0.45]{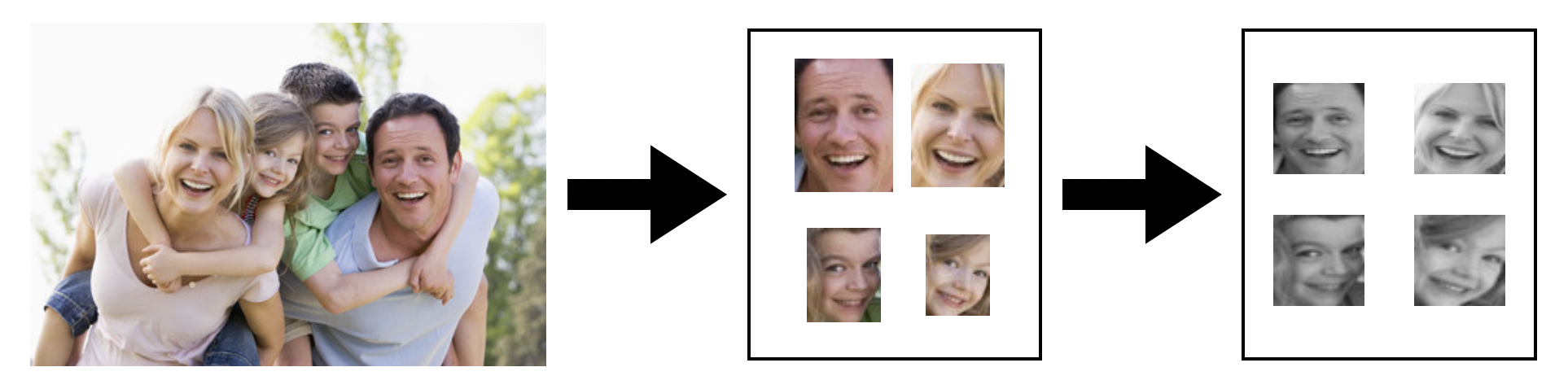}
    \caption{The steps taken to pre-process images from the newly created emotion dataset. First the MTCNN face detector was used to detect all the faces in an image, and crop each one out. Next, simple OpenCV functions \texttt{resize()} and \texttt{cvtColor()} were used to resize the cropped images into $48\times48$ and make them grayscale.}
    \label{fig:preprocess}
\end{figure}

The new emotion dataset contained images with various resolutions and number of people per picture, so in order to combine this with the FER2013 dataset, the faces needed to be cropped from each image. This was done by using a Tensorflow~\cite{tensorflow} implementation of multitask cascaded convolutional networks (MTCNN) for face detection \cite{avrithis2000broadcast} and alignment~\cite{mtcnn}. This code was slightly modified to crop each image around the bounding box detected for each face, and then run on every single image in the collected emotion dataset.
\\ \\
Next, each of the cropped faces were manually inspected to remove any anomalies such as non-faces or re-categorise faces from the background of an image with a different emotion than that of the class it was originally placed in. Due to the vast amount of cropped images, this was a tedious process and prone to human error.
\\ \\
Finally, the OpenCV~\cite{opencv} functions \texttt{resize()} and \texttt{cvtColor()} were applied to each cropped image to ensure they were $48\times48$ in size and grayscale. Figure \ref{fig:preprocess} outlines the brief steps of this pre-processing pipeline. The final images were then combined with the FER2013 images and used in the implementation of the StarGAN, discussed in Chapter \ref{implementation}.

%% file: Sections/8_implementation.tex
For the implementation part of this project, the StarGAN~\cite{stargan} was chosen as it has proven to be very effective with image-to-image emotion generation, and because the authors' PyTorch~\cite{pytorch} code base is surprisingly well commented and easy to understand compared to other official GAN implementations. As mentioned in Chapter \ref{emotion-database}, the emotion dataset compiled throughout this project was not large enough to be used as a training set on its own, so instead it was pre-processed and combined with the FER2013~\cite{fer2013} dataset. The total size of the combined training data was $30,100$ images, split into folders for each of the 7 basic emotions.
\\ \\
The original StarGAN implementation was trained on images of size $256\times256$, so in order for the code to run for the significantly smaller $48\times48$ images a few modifications were made at the command line. The default for the number of convolutional filters in the first layer of the discriminator and generator was $64$, so for this implementation that was decreased to $32$ to avoid a PyTorch division by zero error. The default number of residual blocks in the generator and strided convolutional layers in the discriminator was set to $6$, however this still caused a PyTorch division by zero error so for this implementation the number was varied between $3$ and $5$.
\\ \\
The Imperial Department of Computing GPU cluster (\url{firecrest.doc.ic.ac.uk}) was used to run the jobs for training which significantly decreased training time. On average each experiment took around 2 hours to train, partly due to the small image size, and partly thanks to the GPU cluster.

\section{Analysing Results}
\begin{figure}
    \centering
    \includegraphics[scale=0.5]{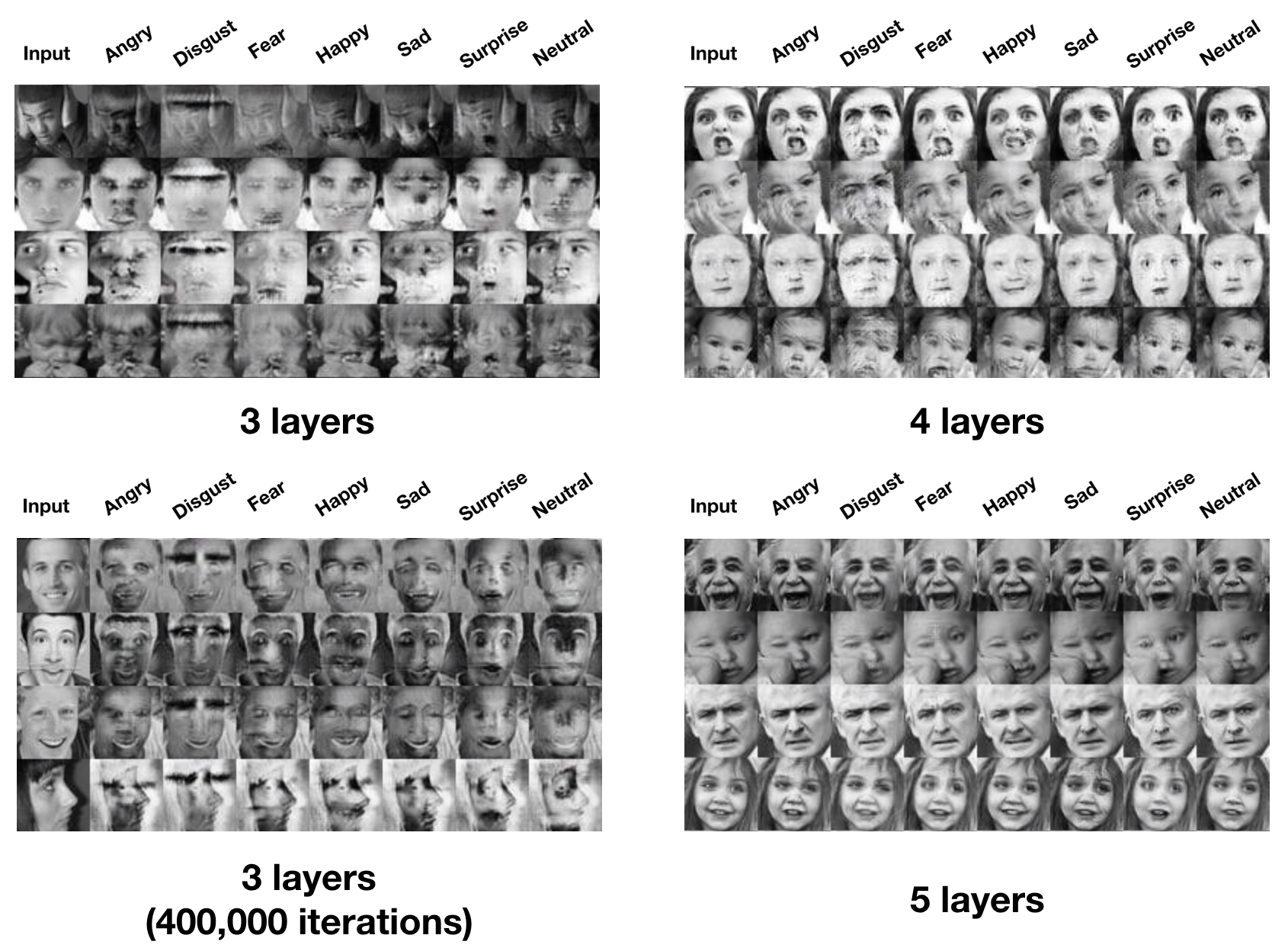}
    \caption{A few of the sample images generated after training the StarGAN with various numbers of layers on the FER2013 dataset combined with the emotion dataset created for this project. The number of layers under each group of images refers to the number of residual blocks in the generator and strided convolutional layers in the discriminator (these are set to the same). Each of these sample images were collected after 200,000 iterations, other than the bottom left one where the images were collected after 400,000 iterations.}
    \label{fig:fer+dataset}
\end{figure}

\begin{figure}
    \centering
    \includegraphics[scale=0.45]{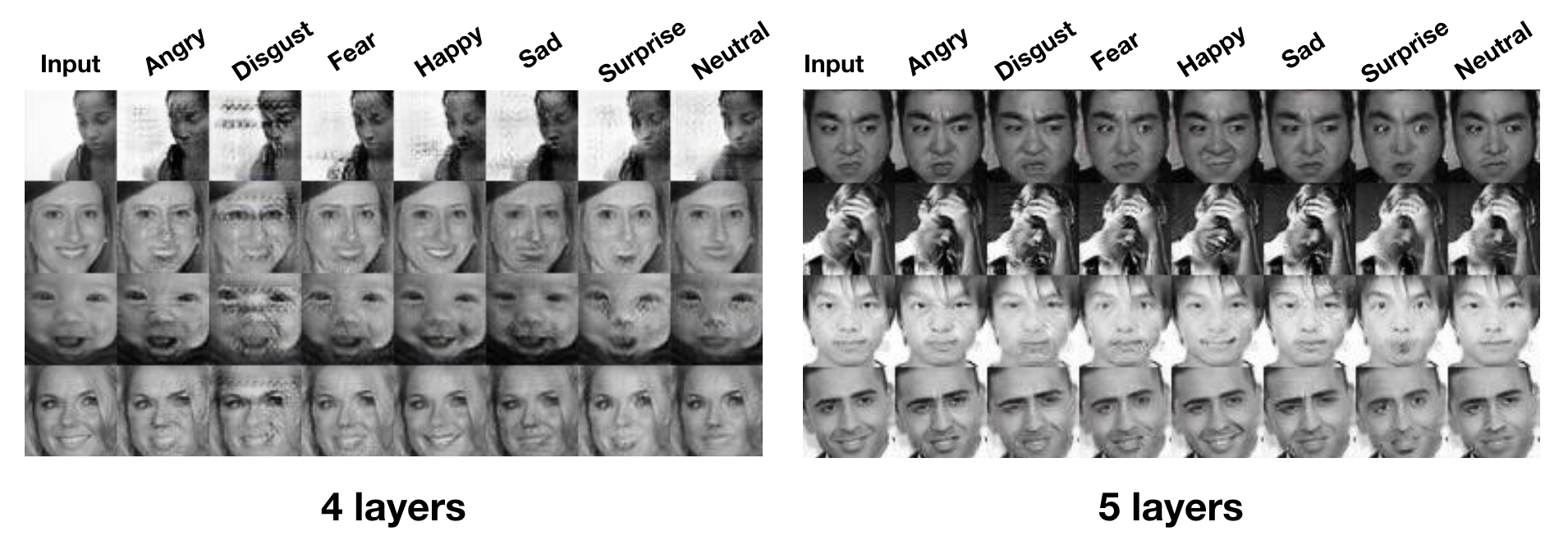}
    \caption{A few of the sample images generated after training the StarGAN with various numbers of layers on the FER2013 dataset. The number of layers under each group of images refers to the number of residual blocks in the generator and strided convolutional layers in the discriminator (these are set to the same). Each of these sample images were collected after 200,000 iterations.}
    \label{fig:fer2013}
\end{figure}

The combined dataset was trained for the default $200,000$ iterations\footnote{An iteration for this implementation is the total number of iterations for training the discriminator, with 5 discriminator updates for every 1 generator update.} for each of the models in this experiment. It was trained on models with 3, 4, and 5 layers\footnote{A layer here refers to the number of residual blocks in the generator and the number of strided convolutional layers in the discriminator.}, and a few of the resulting sample images generated can be seen in Figure \ref{fig:fer+dataset}. Due to the bad results obtained from the 3 layer model it was run for 200,000 iterations longer to see if extending the training time would help the model converge. All this accomplished, however, was creating a clearer face outline in the generated image, but it still did not manage to generate photo-realistic results.
\\ \\
Looking at the generated sample images it becomes evident that the greater the number of layers of the model the more photo-realistic the generated images are. However, although the images generated by the 5 layer model are the most realistic, very few actually resemble the various emotions they are aiming to portray, and instead all 7 images look almost identical. The third row of the 5 layer model comes the closest to matching the emotions with a clear change to the mouth for happy and eyes/mouth for surprised. Overall though, none of the attempted models for this manage to generate images that are both photo-realistic and reflect each of the 7 emotions.
\\ \\
Although the images are not particularly impressive, they do show that the StarGAN has indeed learned things about the 7 basic emotions and certain characteristics they have in common. Even for the quite terrifying and non-realistic faces generated by the 3 layer model there is a clear o-shaped mouth added to all the surprised images indicating that the GAN has learned that the most important feature indicating surprise is that the mouth makes an o-shape.
\\ \\
A reason why the GAN seems to struggle with generating images could partly be due to the `in the wild' nature of them. When the GAN with such limited training data is trying to learn important features indicating each emotion and figure out which part of the face is the mouth/nose/eyes simultaneously, its performance suffers. The fact that many images show a hand over the mouth, or a face turned sideways makes it nearly impossible for the GAN to even locate where the facial features \cite{kollias7,kollias14} are for it to modify, resulting in the non-facelike images generated by the 3 layer model.
\\ \\
Another reason why this GAN struggles with training is most likely due to the image size. When the images are smaller, specific details become harder to notice in an image as a single pixel represents more than it would if the same image was larger. This means that many of the fine details regarding how a mouth looks at the corners or how eyes are placed and shaped on a persons face become a lot harder for the GAN to learn. This makes it almost impossible for the GAN to successfully generate images that look human-like as it has not been able to properly learn what that looks like.
\\ \\
An obvious obstacle is also the fact that these models are trained on a combination of two datasets, one much smaller than the other and possibly acting more as noise \cite{raftopoulos2018beneficial} than as a contribution to the larger one (FER2013). Due to this the better models (4 and 5 layers) were also trained on the FER2013 dataset alone (see Figure \ref{fig:fer2013}).
\\ \\
As the sample images show, when the models were trained on solely FER2013 they performed much better. The generated images are still far from perfect, but looking at the first row of generated images for the 5 layer model shows 7 photo-realistic images with decent resemblance to their conditioned emotions. The input image on the first row (of the 5 layer model) definitely makes the generation task easier by not having the face turned sideways or having anything else to distract in the image (unlike the image in row 2 where the face is rotated and a hand covers part of it, and as a result none of the emotions are evident). This confirms the suspicion that combining the two datasets made the new dataset act more like noise than a helpful addition to the training data, hindering the model from learning crucial features of each emotion.
\\ \\
Another clear observation is that increasing the number of layers improves the overall performance. Due to the small image size the layers cannot be increased much further without reducing the filter size in the first layer of the discriminator and generator. If future work were to be conducted on this project it would be interesting to see the effects of decreasing the filter size and increasing the number of layers, and whether this would continue the trend of improving the overall performance. It would also be interesting to increase the overall image size to $96\times96$ or $128\times128$ as this would most likely help improve performance as well.
\clearpage

%% file: Sections/9_conclusion.tex
The main goal of this project was to undertake a survey of GAN research to date and determine possible areas for future research. This was discussed thoroughly in Chapter \ref{gan-discussion}, with the main suggestions for future work being:
\begin{itemize}
    \item \textbf{Evaluating GANs} - there is still no one recommended method for evaluating the overall performance of GANs, making it increasingly difficult to compare the performance between various architectures.
    \item \textbf{Understanding GANs} - how and why GANs work and what they actually learn remains an open question. Although some questions were answered by Bau et al.~\cite{gan-dissection}, many still remain.
    \item \textbf{Techniques for training GANs} - although some progress has been made with regards to training GANs in a more stable manner (for example by introducing the Wasserstein distance as a loss function~\cite{wgan}), stability in training remains a large problem for GANs. This was amplified by recent proposals such as the BigGAN~\cite{big-gan} where scaling up introduced numerous training instability issues.
\end{itemize}

These areas for future work suggest that the better we can understand how GANs work, and comparatively measure them against each other, the better future GAN models will perform.
\\ \\
The second part of this project entailed compiling a dataset of emotion images `in the wild' of all the 7 basic emotions. This was done successfully with a resulting database of $1,463$ images. These images were combined with the FER2013 images and used to train a StarGAN to generate emotions on images of faces. As the results from this implementation (Chapter \ref{implementation}) showed this was not particularly successful. Although the StarGAN seemed to learn certain features about the emotions, such as an o-shaped mouth being common for surprise, it was often unable to generate photo-realistic results of the various emotions. 
\\ \\
This was partly to do with the emotion dataset created during this project acting more as noise than a helpful addition to the FER2013 dataset, and partly due to the images being too small. The size of the images meant that each pixel contained more details than it would in a larger image, making it much more difficult for the GAN to distinguish smaller details in the image and thereby recreate them in a realistic way. When training the StarGAN on only the FER2013 dataset this helped solve the issue of the new dataset acting as noise, but as the image size remained the same, it was still not able to learn the finer details of the images.